\def\year{2022}\relax
\documentclass[letterpaper]{article} 
\usepackage{aaai22}  
\usepackage{times}  
\usepackage{helvet}  
\usepackage{courier}  
\usepackage[hyphens]{url}  
\usepackage{graphicx} 
\usepackage{natbib}  
\usepackage{caption} 
\DeclareCaptionStyle{ruled}{labelfont=normalfont,labelsep=colon,strut=off} 
\usepackage{booktabs}
\usepackage{adjustbox}
\usepackage{todonotes}
\usepackage{multirow}
\usepackage{amsmath}
\usepackage{amssymb}
\usepackage{newfloat}
\usepackage{listings}
\nocopyright
\title{Can Open Domain QA Models Answer Visual Knowledge Questions?}
\author{
    Jiawen Zhang \qquad Abhijit Mishra \qquad Avinesh P.V.S\\
    Siddharth Patwardhan \qquad Sachin Agarwal
}
\affiliations{
    Apple\\
    1 Apple Park Way, Cupertino, CA 95014\\
    \texttt{\{jiawen\_zhang2, abhijitmishra, avineshpvs, patwardhan.s, sachin\_agarwal\}@apple.com}
}

\begin{document}

\maketitle

\begin{abstract}
The task of {\em Outside Knowledge Visual Question Answering} (OKVQA) requires an automatic system to answer natural language questions about pictures and images using external knowledge.
We observe that many visual questions, which contain deictic referential phrases referring to entities in the image, can be rewritten as ``non-grounded'' questions and can be answered by existing {\em text-based} question answering systems.
This allows for the reuse of existing text-based Open Domain QA Systems for {\bf visual} question answering.
In this work, we propose a potentially data-efficient approach that reuses existing systems for (a) image analysis, (b) question rewriting, and (c) text-based question answering to answer such visual questions.
Given an image and a question pertaining to that image (a {\em visual question}), we first extract the entities present in the image using pre-trained object and scene classifiers.
Using these detected entities, the visual questions can be rewritten so as to be answerable by open domain QA systems.
We explore two rewriting strategies: (1) an unsupervised method using BERT \cite{devlin2018bert} for masking and rewriting, and (2) a weakly supervised approach that combines adaptive rewriting and reinforcement learning techniques to use the implicit feedback from the QA system.
We test our strategies on the publicly available OKVQA dataset \cite{okvqa} and obtain a competitive performance with state-of-the-art models while using only 10\% of the training data.
\end{abstract}

\section{Introduction}
\label{sec:intro}
Within the realm of question answering systems, {\em Visual Question Answering} (VQA) seeks to answer questions pertaining to given pictures or images.
Broadly, VQA can be categorized into three types --- (a) \textit{direct} (e.g., ``How many slices of pizza are there?''),  (b) \textit{visual common sense reasoning} (e.g., ``Why is the person running?''), (c) \textit{outside knowledge} (OK) (e.g., ``Where does this food originate from?'').
In this work, we focus on studying ``outside knowledge'' queries, which require external knowledge from either free text documents (e.g., Wikipedia), knowledge graphs (e.g., DBpedia), or large language models (e.g., BERT) to answer questions about contents of the image. 

Existing approaches on OKVQA tackle this task by combining text and image representations for a joint QA system \citep{vl-bert, ViLBERT}. 
However, they requires large amounts of high-quality multi-modal training data containing images, a question to each image that draws upon external knowledge, and some human-annotated answers for each question.
Such high-quality data is often harder to collect \cite{okvqa} than individual datasets used for well-known problems in computer vision and NLP (e.g., entity extraction from images, text-based QA).
Furthermore, according to \citet{okvqas3}, a surprisingly large fraction of the queries in the available OKVQA dataset \citep{okvqa} does not need external knowledge to answer.
Instead, some are independent of the image, some depend on speculation (e.g., ``Can you guess what it is?''), some require object recognition (e.g. ``What type of plant is it?'') or are otherwise answerable from the image alone.
To simplify human annotation requirements, the OKVQA dataset treats the task as a classification task, selecting from among the most frequent answers in the training set rather than obtaining the answers from an external source (such as, a knowledge graph or spans from a document corpus like Wikipedia).

\begin{table}[ht]
\footnotesize
\begin{centering}
\begin{tabular}{l|p{5.5cm}}
\toprule
    Question & How tall is this animal on average? \\
    \midrule
    (1) Objects & giraffe, stone, tree, park\\
    (1) Rewrite & How tall is \textbf{giraffe} on average?\\
    (2) TextQA & 15 feet \\
\bottomrule
\end{tabular}
\end{centering}
\end{table}

\begin{figure}[ht]
\centering
\includegraphics[width=0.8\linewidth, height=140 pt]{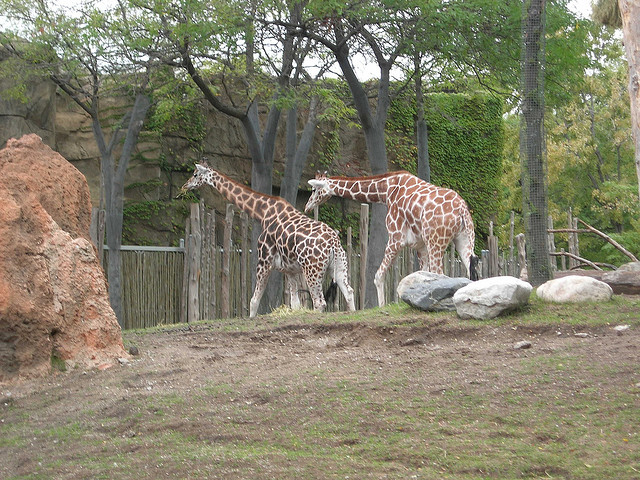}
\caption{Example OKVQA image and question describing steps needed to solve using a text-based model}
\label{fig:intro}
\end{figure}

Since several pre-trained models are already available for image analysis \cite{he2016deep,iandola2016squeezenet,wu2019detectron2} and for text-based question answering \cite{2020t5cqba,lewis2020retrieval,chen2017reading}, a potential solution to mitigate the data scarcity problem is to leverage such pre-trained models.
We propose a potentially data-efficient approach that reuses existing systems for (a) image analysis, (b) question rewriting, and (c) text-based question answering to answer such visual questions.

Given an image and a question pertaining to that image (a {\em visual question}), we first extract the entities present in the image using pre-trained object and scene classifiers.
Using these detected entities, the visual questions can be rewritten so as to be answerable by open domain QA systems.
We explore two rewriting strategies using unsupervised and weakly supervised techniques that do not require any additional data for training a rewriter.

In the unsupervised approach, given a question and a list of entities as input, we find all possible replacements of an entity with a span and leverage existing language models like BERT \cite{devlin2018bert} and GPT-3 \cite{gpt3} to select the semantically correct entity.
While this method solely depends on the pre-trained language models' syntactic ability to choose the best rewrite, the reformulated questions may not necessarily follow the same vocabulary and syntactic structure that is seen by the text-based QA model during training.
Thus, for the second rewriting strategy, our goal is to adapt to the input style of the text QA model.
In this, we train a sequence-to-sequence-based rewriter such as BART \cite{lewis2019bart} implicitly, without the need of a labeled dataset of original and rewritten questions, with the help of reinforcement learning (RL).
We leverage feedback from the QA system and correctness of the answer candidates to reward the RL.

We test the above two strategies on a portion of the publicly available OKVQA dataset \cite{okvqa} that contains only knowledge-oriented visual questions.
We reuse T5-based closed-book question answering system \cite{2020t5cqba} as our text-based model and BART as our rewriter model for implementing the two strategies mentioned above.
We choose the T5 model purely because it is easy to deploy, does not depend on external resources or models (such as \textbf{retrievers} in other pipelines), and delivers state-of-the-art performance on knowledge-oriented question answering datasets.
Furthermore, experimental results reveal that our proposed rewrite-based pipelines have an exact match and semantic-similarity-based scores competitive to the other end-to-end VQA models.

The primary contributions of this work are the following:
\begin{itemize}
    \item We explore two rewriting strategies, a QA model agnostic (unsupervised) and a QA model-driven (weakly supervised), to transform deictic knowledge-oriented visual questions into direct questions better understood by text-based QA models.
    \item Unlike end-to-end outside knowledge VQA models that try to capture images, questions, and external knowledge through a single model with a limited number of parameters, our approach is modular.
    As a result, it can reuse richer state-of-the-art individual models for image label extraction, rewriting, and question answering.
    \item Our QA model-based strategy uses an architecture of adaptive rewriting with reinforcement learning techniques using fractionally lesser data (1/10$^{th}$) compared to state-of-the-art systems.
\end{itemize}

\section{Related Work}
\label{sec:related}
In this section, we discuss related work to our research concerning:
(1) Text-based QA Models, (2) Visual Question Answering, (3) Adaptive rewriting techniques, and (4) Reinforcement learning for rewriting. 

\subsection{Text-based QA Models}

Open-domain question answering has been predominantly based on the \textbf{retrieve-then-read} based paradigm \cite{zhu2021retrieving}.
Excerpts from large-scale web documents (e.g., Wikipedia) or knowledge representations from massive KGs are first retrieved and then comprehended by machine-learned models to yield potential answers from the extracted passages. 
Knowledge-oriented question answering, a type of open-domain question answering, often draws on two types of information \cite{fu2020survey}.
The first relies on retrieving many documents from a structured or unstructured knowledge source, such as Wikipedia.
The response is then extracted by conditioning on the query and the retrieved passages.
REALM \citep{guu2020realm} and ORQA \citep{lee2019latent}, for example, have demonstrated promising results when combining masked language models with a differential retriever.
The second type relies on implicit knowledge stored in model parameters.
For example, T5 \citep{2020t5cqba} distributes knowledge in its parameters in a possibly inexplicable way using a large language model pre-trained on unstructured text, which delivers well on knowledge-oriented questions without having to mine knowledge from external sources.
There have also been hybrid models that combine parametric and non-parametric (i.e., retrieval-based) memories, allowing information to be assessed, changed, and expanded.
For example, \textit{Retrieval-Augmented-Generation} (RAG) \citep{lewis2020retrieval}, enables it to update internal knowledge based on newly obtained data.
\begin{figure*}[tb]
  \includegraphics[width=500 pt,height=110 pt]{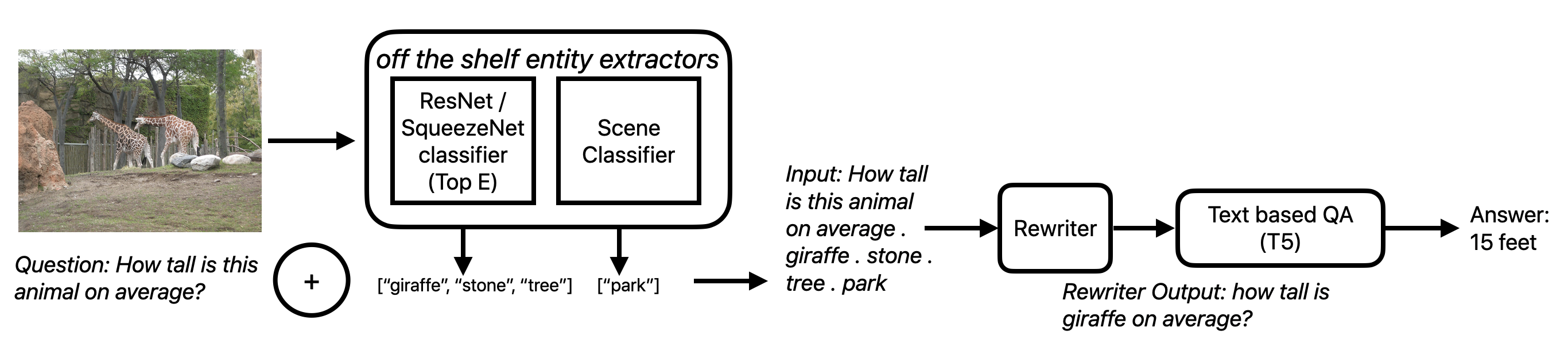}
  \caption{End-to-end flow with entity extraction and input adaptation}
  \label{fig:end-to-end}
\end{figure*}
\subsection{Visual Question Answering}
A growing interest in multi-modal research and the advent of high-capacity deep neural learning models has made visual question answering a trending research topic. 
\citet{wu2017visual} provide a summary of VQA models and datasets, where they focus on deep learning models that perform reasonably well on public datasets such as VQA \cite{antol2015vqa} and OKVQA \cite{okvqa}.
Early approaches to VQA combined recurrent networks with CNNs to integrate textual and visual data, transformer-based models to highlight the most relevant image region \citep{Ben-younes_2017_ICCV,wang2020vd}, and zero-shot learning to counter the lack of labeled examples \cite{demirel2019image}.
None of these approaches, however, are built to use external knowledge, thus they can't manage situations where the image doesn't reflect all of the information needed to answer the query. 
Some research studies \citep{cao2021knowledge, ziaeefard2020towards} have tackled knowledge-based visual queries, but they exclusively deal with structured knowledge sources represented as subject-relation-object or visual concept-relation-attribute triplets.
Therefore, the capacity to answer open-ended and complex questions is limited.

Most relevant to us, \citet{okvqas3} have proposed to treat OKVQA as a task of fusing structured data from the image with the unstructured text rather than a visual recognition problem.
The idea is to transform the multi-modal input (image + text) to a text-only input so that the text-based QA model can directly interpret and answer (Figure~\ref{fig:intro} shows a sample).
The text-only version of the original question is constructed by: (a) \textbf{Selecting} a referential expression (e.g., ``this animal''), (b) \textbf{Substituting} the expression with the most appropriate entity from the image, and (c) \textbf{Searching} and augmenting additional knowledge with the help of a search engine.
For selection and substitution, \citet{okvqas3} use supervised learning approaches for predicting referential spans and selecting appropriate entities for substitution.
This approach, however, would be harder to scale OKVQA systems to unseen domains and question types.
Moreover, the output of such a rewriter may not always align with the input vocabulary of the text-based QA models used for answering.
In our work, we emphasize using unsupervised and weakly supervised methods that extract grammatical and contextual cues from the visual questions to reformulate them.

\subsection{Adaptive Rewriting Techniques}
Since fine-tuning the large pre-trained model could be expensive if there are many downstream tasks, \citet{houlsby2019parameter} propose a new transfer mechanism: adding only a few trainable parameters with adapter module per task so that new tasks can be added without revisiting previous ones.
Adapters serve the same purpose as fine-tuning but do it by stitching in layers to the main pre-trained model, and updating the weights of these new layers while freezing the weights of the pre-trained model.
Seq2Seq based adapters (rewriters) are not new and have been applied to tasks like accent adaptation \citep{khandelwal2020black} and response generation \citep{madotto2020adapter}, but not for adapting to knowledge-based visual QA.
Based on the idea of adapter, \citet{pfeiffer2020adapterhub} introduces Adapterhub\footnote{https://adapterhub.ml}, a new framework built on top of the HuggingFace library.
These adapter architectures adapt well to tasks based on top of generic encoders like BERT, but they can not easily adapt to different input styles of QA models.

\subsection{Reinforcement Learning for rewriting}
Reinforcement learning has been successfully applied to various NLP tasks, including summarization \citep{DBLP:conf/iclr/PaulusXS18}, question answering \citep{hua-etal-2020-shot}, and many more.
Only recently, Reinforcement Learning (RL) has been applied for query rewriting.
\citet{buck2017ask} propose an approach to rewrite a jeopardy style query (declarative) to a question answering system that is often trained on questions (interrogative).
The reformulation system is trained end-to-end to maximize answer quality using policy gradient.
The policy gradient approach has also been used for tasks like machine translation \citep{wu2018study}, which has shown decent performance improvement even when less labeled data is available.
\begin{figure*}[htb]
  \includegraphics[width=\textwidth,height=190 pt]{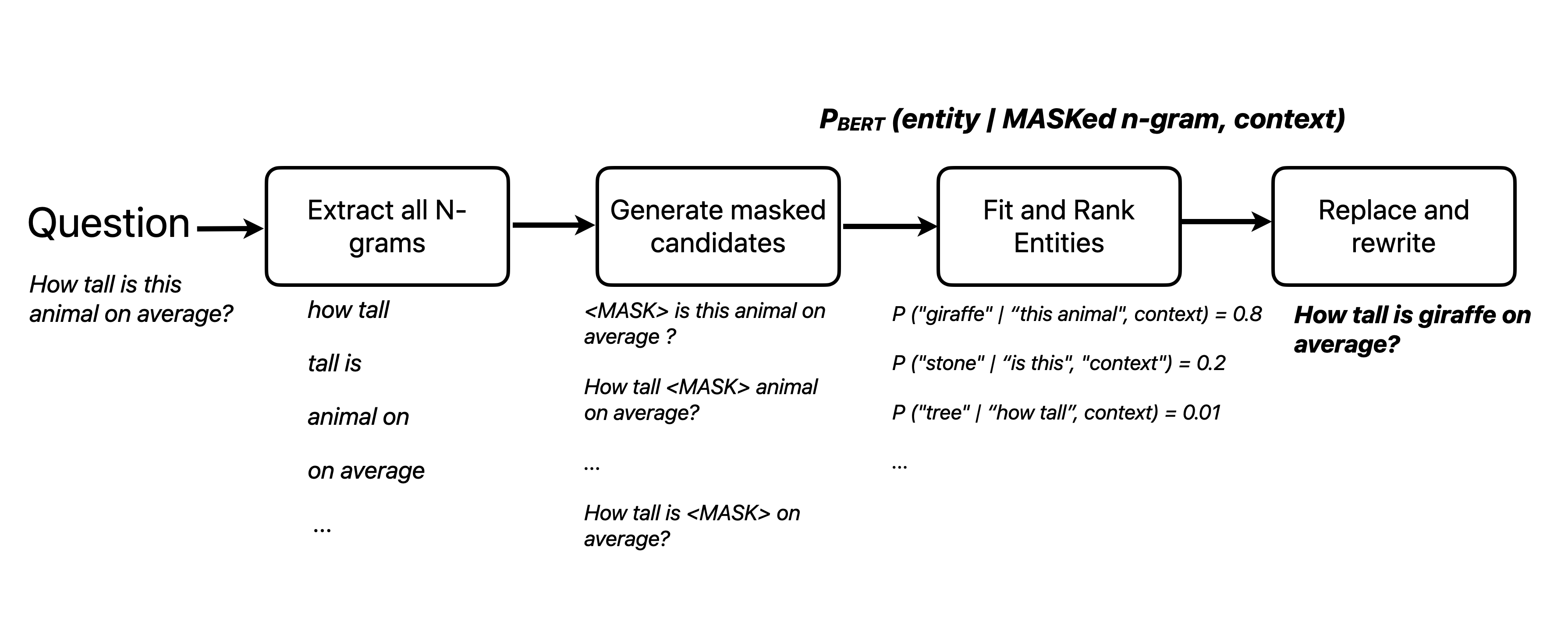}
  \caption{QA Model Agnostic Rewriting with Mask-Replace-Rank based approach using BERT}
  \label{fig:mask-fit}
\end{figure*}
\section{Method}
\label{sec:method}
Inspired by the idea of adapter \citep{houlsby2019parameter}, our solution is to adaptively rewrite the query and produce a "non-grounded" question, which can be answered by the \underline{text-based} QA system.
As shown in Figure \ref{fig:end-to-end}, given an input question (e.g., {\em How tall is this animal on average?}) and an image of two giraffes, we first extract the top $E$ entities from the image using off-the-shelf entity extractors such as SqueezeNet \citep{iandola2016squeezenet} for object entities and Scene Classifier \citep{zhou2017places} for scene labels.
Those entities are then provided as input to a rewriter along with the original question with the objective of rewriting the question into an independent form.
We take a simple approach of concatenating the entities with the question (e.g. {\em how tall is this animal on average . giraffe . stone . tree . park}), and allowing the model to learn the rewrite transformations.
Our adaptive rewriting model is a generative model that learns to rewrite this input into an ``independent" question (e.g. {\em how tall is giraffe on average?}).

The rewritten question is then passed on to the text-based QA model, to obtain an answer.
Adaptive rewriting offers the following advantages.
(1) since we directly use an off-the-shelf QA system, the core QA model does not have to be retrained on this kind of input and is allowed to remain agnostic of the changes in the input structure.
(2) Adapter rewriting module can be added/removed from the execution pipelines on-demand.
This allows for a modular system that is easy to debug and analyze.
(3) the predicted answer is more interpretable, since we can see the rewritten that resulted in the final answer. 

As mentioned in Section \ref{sec:intro}, we explore two approaches for adaptive rewriting: {\em QA Model Agnostic Rewriting} and {\em QA Model Aware Rewriting}.
We reiterate that these approaches do not require additional labeled training data of original and rewritten questions.

\subsection{QA Model Agnostic Adaptive Rewriting}
\label{sec:agnostic-rewrite}
For model agnostic rewriting, we implement a fairly straightforward \textbf{Mask-Replace-Rank} approach \cite{havens2019fitbert}.
As shown in Figure \ref{fig:mask-fit}, given a question ``{\em how tall is this animal on average?}'', we extract all N-grams ($N={1,2,3}$) from the question, and generate the masked candidates by replacing the N-grams with the $\mathrm{<MASK>}$ tokens.
For each entity extracted from the image, a pre-trained BERT model (BERT-large \cite{devlin2018bert}) is used to compute the probability of replacing the mask with the entity ($P_{bert}$(entity $|$ MASKed  N-grams, context)).
Entities are ranked according to the language model probability of the token sequence and the top scoring token sequence is selected as the rewrite.
In our example, $P$(giraffe $|$ this animal, context) returns the highest score since language model prefers the sequence ``{\em how tall is giraffe on average?}'' to other candidate sequences.
This approach leverages the fill-in-the-blanks nature of the BERT model without requiring additional training data.

\subsection{QA Model Aware Adaptive Rewriting}
\label{sec:aware-rewrite}
While the prior approach relies primarily on identifying the more probable token sequences for that language, we expect that including some signal from the question answering task would help the model sort out the more ambiguous cases.
In this approach, the model-aware method uses the gold-standard answers and the QA model to learn to rewrite.
We use a generative sequence-to-sequence rewriter that takes inputs over vocabulary $V_{0}$ and produces an output sequence over vocabulary $V_{1}$, which in practice is the same as the vocabulary of the black-box QA model.
To begin with, this rewriter can be any generic, pre-trained sequence to sequence model (such as BART), and is fine-tuned with noisy inputs \cite{morris-etal-2020-textattack} that simulate our use-case scenarios. 

The difference between a traditional sequence-to-sequence model and the rewriter is that the rewriter is not retrained with direct supervision of gold-standard sentences provided as output labels.
It is rather fine-tuned based on the learning signals (gradients) from the black-box text-based QA model. 
This way of training would not only help train the rewriter train with less data, but it will also enable it to adapt to different QA systems during training.
Since most of the submodules in the architecture are frozen, the rewriter should learn faster and mitigate the problem of catastrophic forgetting during adaptation.

\begin{figure*}
   \includegraphics[width=\textwidth]{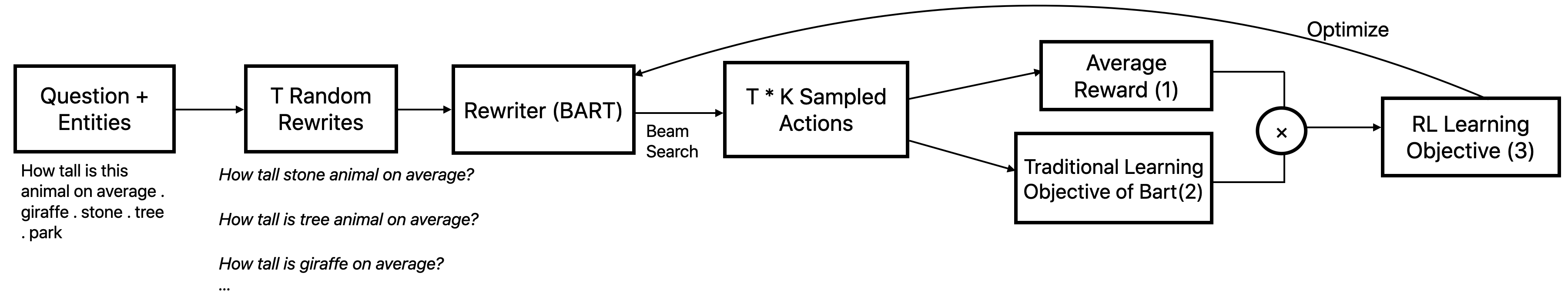}
   \caption{End-to-end flow with Policy Gradient Optimization}
   \label{fig:rl-steps}
 \end{figure*}
We use BART \cite{lewis2019bart} for our rewriter, keeping its input and output vocabulary shared and same as the text-based question answering model (T5).
The rewriter is initialized with a pre-trained version of BART trained on a large corpus with a denoising auto-encoding objective.
The model then undergoes a reinforcement learning process, comprised of three steps: (a) Exploration (b) Reward Computation (c) Optimization (Figure \ref{fig:rl-steps}).

\subsubsection{Exploration:}
In this step we allow the rewriter to produce a set of candidates.
In reinforcement learning settings for sequence-to-sequence networks, beam-search decoding is often employed to generate candidates.
Our BART decoder generates top $K$ candidates ($K=3$) through beam search with a beam width of $5$.
Initially, the pre-trained version of BART (which is trained with an auto-encoding objective) may not offer a lot of variations with beam search and may end up copying the input.
So, to allow some syntactic variations, we consider the following strategy:
\begin{enumerate}
\item We prepare several temporary inputs following a similar method as Section \ref{sec:agnostic-rewrite}). 
We first extract all N-grams (e.g., $N={1,2,3}$) from the question and replace them with entities from the image.
For each entity, we filter the top $T$ candidates (in our work, $T = 10$) using a GPT-2 language model \cite{radford2019language}, using the likelihood (log probability) of the rewrites.
\item We then pass these $T$ temporary rewrites through the BART encoder-decoder, which generates $K$ candidates for each temporary rewrite.
For one entity the total number of candidates is $T*K$.  
\item For each entity, we compute the average reward of the generated $T*K$ candidates following the steps in Section \ref{sec:reward_comp}. 
or optimization, we only use the candidate lot that has the maximum average reward.
\end{enumerate}

\subsubsection{Reward Computation:}
\label{sec:reward_comp}
Each candidate output is given to the QA model which predicts the candidate answer.
For $T*K$ candidates we compute the average rewards ($\mathcal{R}$) as follows:
\begin{equation}
\mathcal{R} = \frac{1}{|S|} \sum_{i = 1}^{|S|} cos (e(y), e(\hat{y_{i}}))
\label{eq:reward}
\end{equation}
where, $S$ represents the $T*K$ candidates, $y$ and $\hat{y}$ are the gold standard and predicted answers respectively, $e(.)$ is the embedding representation of the answers, and $cos(.)$ is the cosine similarity between two embeddings.
As the answers may have multiple words, for embedding extraction, we consider a sentence embedder such as BERT. 

The rationale behind choosing this reward function is two-fold: (a) it is continuous and offers smooth policy gradients (a) it indicates the degree with which the predicted answers are semantically similar to the gold-standard answers.
With this design, the rewriter output ``how stone this animal on average" should receive low rewards, as it does not have adequate information to obtain a correct answer.

\subsubsection{Optimization:}
\label{sec:optimization}
In the optimization step, the rewriter model learns to exploit the candidates that maximize the average reward ($\mathcal{R}$).
In a traditional setting for BART, for $B$ training instances of inputs (concatenated question and entities) and rewrites, $\{x^i,r^i\}_{i=1}^{B}$, the learning objective (L) is to maximize the likelihood of a target $r^i$ given ${x^i}$, which is similar to minimizing the \textit{cross entropy} between the target distribution and the predicted output distribution.
For training the network, the gradient of the negative cross-entropy loss is considered to update the model parameters.
The gradient is given by:
\begin{equation}
\nabla_\theta L(\theta) = \nabla_\theta \sum_{i=1}^{B} r^i log~ P_{M_\theta}(\hat{r^i}|x^i)
\label{eq:normalgradient}    
\end{equation}
where $M_\theta$ is the BART model with parameters $\theta$. 
In our  reinforcement learning setting, for optimizing the reward that comes from an external source, we use the policy gradient mechanism \cite{williams1992simple}.
The generator of BART, operating with a policy of $P_{M_\theta}(\hat{r^i}|x^i)$, producing an output $\hat{r^i}$ with an \textit{expected} reward ($\mathcal{R}$) computed using Equation \ref{eq:reward}, will thus have the following gradient:
\begin{equation}
\begin{split}
    \nabla_\theta RL(\theta) & = \nabla_\theta\mathbb{E}_{\hat{r_i}\sim P_{M_\theta}(\hat{r^i}|x^i)}[\mathcal{R}_{r^i}] \\
    & = \mathbb{E} [\mathcal{R}_{r^i})\nabla_\theta log(P_{M_\theta}(\hat{r^i}|x^i))]
\end{split}
\label{eq:policy}
\end{equation}
where $RL$ is the modified learning objective which has to be maximized.
This way of optimizing ensures that at the end of the training, the BART model learns to translate the concatenated input to a rewritten by dropping entities and performing necessary re-ordering.
For implementation, we use cross-entropy loss for the policy in Equation \ref{eq:policy}, by computing the cross entropy between the candidates and the input, using the input as reference.
This helps in penalizing too many edits (e.g., adding or dropping too many terms in the rewrite).

\section{Experimental Setup}
\label{sec:setup}

In this section, we describe the setup for our experiments, including the dataset, evaluation metrics, baselines and training details that we will use to assess the effectiveness our approach.

\subsection{Dataset}
We test our method on the OKVQA$_{S3}$ \citep{okvqas3} dataset, which annotates four types of questions in the OKVQA: (1) question which require detecting objects and subsequent reasoning over an external knowledge source to arrive at the answer, (2) question which require reading text from the image (OCR) (and no other information) to answer, (3) question which are based on personal opinion or speculation, (4) Other. 
Since we focus on rewriting knowledge-based queries, we take only the questions annotated as the first type, which have 1,643 questions in the training dataset and 999 questions in the test dataset. We further deleted the questions that require spatial reasoning (e.g., \textit{the person on the left}).
After the data processing work, we eventually have 1,010 training examples and 363 test examples (Table \ref{table:datasets}). 

\begin{table}[h]
\centering
\footnotesize
\begin{tabular}{lccc}
\toprule 
Dataset & \#Train & \#Test \\
\midrule 
OKVQA &9,009 & 5,046 \\ 
OKVQA$_{S3}$ &1,643 & 999 \\
Our Final Dataset &1,010 & 363 \\ 
\bottomrule 
\end{tabular}
\caption{Statistics of the Knowledge VQA Datasets}
\vspace{-0.6cm}
\label{table:datasets}
\end{table}

\subsection{Systems for Comparision}
We compare our approach with five existing VQA methods\footnote{We obtained predictions for OKVQA \cite{okvqa} from the author} along with two additional baselines: one uses a concatenated version (without rewrite) as input to the T5 model and the other fine-tunes T5 with those concatenated inputs. The VQA baselines are:
\begin{itemize}
    \item MUTAN \citep{ben2017mutan}: A tensor-based Tucker decomposition to efficiently parametrize bilinear interactions between visual and textual representations.
    \item MUTAN+AN \citep{okvqa}: An attention-based version of MUTAN that fuses MUTAN and ArticleNet (MUTAN + AN) as a knowledge-base baseline. 
    \item BAN \citep{kim2018bilinear}: Bilinear Attention Networks that uses a co-attention mechanism between the question and image features.
    \item BAN+AN \citep{okvqa}: An attention-version of BAN that concatenates the output of the memory network with the last BAN hidden state.
    \item QOnly \citep{okvqa}: An neural network with 3 hidden layers that only takes question features.
\end{itemize}

\subsection{Evaluation Metrics}
To evaluate the QA system, we employ three widely used metrics for QA systems (a) Exact match (EM), (b) Bert Similarity Score (BS), (c) Human evaluation (HE)

\begin{itemize}
    \item \textbf{EM:} We calculate the exact match score between the golden answers and the predicted answers, and take the average across all answers.
    \item \textbf{BS:} We use SentenceBERT to calculate the similarity between the golden answers and the predicted answers and take the average score across all answers.
    \item \textbf{HE:} We manually evaluate the quality of the answers and based on the question and the image. The result should be either correct or incorrect. We asked four annotators to help us grade the 363 test data for all baselines. They are provided with a UI and guidelines (Section \ref{sec:guidelines}). 
\end{itemize}



\subsection{Grading Guidelines for Human Evaluation}
\label{sec:guidelines}
There are 363 examples to be graded. For each system output, the human annotator marks whether the answer and rewritten output given by the system are correct or not (binary). Each example contains an image, a question about the image, a list of gold standard answers from human annotators, and answers from eight Question Answering systems. While evaluating the answers, note that for certain questions multiple answers are possible. Moreover, some of the answers may not appear in the gold standard answer list. So, in some cases they mark correct vs. incorrect according to their best judgment. If required, they are allowed to look up the question on a Web search engine. Certain questions require the answers to be ``numbers" (such as calorie content in food or year of invention, etc.). The predicted answer is considered to be correct if it is within a small window of the correct answer (e.g., 19th century for 1890).

\subsection{Model and Training Details}
For BERT-based scoring and ranking, we use FitBERT\footnote{https://github.com/writerai/fitbert} and GPT based scoring is done through LMScorer\footnote{https://github.com/simonepri/lm-scorer}. The model optimization was carried out with a batch size of 16 for 9,600 steps.
We allocated 1 GPU and 32 CPUs for all of our experiments, and on average for each experiment, we used 8 percent of the allocated GPU, 60 percent of the allocated CPUs, and 35 percent of the GPU memory.

\section{Results and Analysis}
\label{sec:results}


\begin{table}
\centering
\footnotesize
 \begin{tabular}{l | c| c | c | c} 
 \toprule
 Models & Train Data & EM & BS & HE\\ [0.2ex] 
\midrule
\multicolumn{4}{l}{\textit{VQA Models}} \\[.2ex]
  MUTAN+AN & 9,009 & 0.28 & 0.70 & 0.45 \\
  BAN  & 9,009 & 0.29 & 0.70 & 0.47 \\
  BAN+AN &9,009 & 0.29 & 0.70 & 0.43 \\
  MUTAN & 9,009 & 0.30 & 0.69 & 0.43 \\
  QOnly &9,009 & 0.16 & 0.62 & 0.24 \\
   \midrule
\multicolumn{4}{l}{\textit{Baselines}} \\[.2ex]
Concatenated Input & - & \textbf{0.32} & \textbf{0.71} & 0.54 \\
 Fine-Tuned T5 &9,009 & 0.30 & \textbf{0.71} & 0.48\\

  \midrule
 \multicolumn{4}{l}{\textit{Our Methods}} \\[.2ex]
 Model Agnostic & - & 0.31 & 0.70 & \textbf{0.67} \\
 Model Aware &1,010 & 0.29 & 0.69 & \textbf{0.67}\\
\bottomrule
\end{tabular}
\caption{Evaluation of answers and rewrites using three different metrics: EM: Exact Match, BS: Bert Similarity, HE: Human Evaluation}
\label{tab:results}
\end{table}

\begin{table*}[t]
    \centering
    \footnotesize
    \resizebox{2.1\columnwidth}{!}{
    \begin{tabular}{p{4.5cm}|p{2.5cm}|p{2.8cm}|l|p{5cm}|l}

\toprule
\bf Image & \bf Question & \bf Extracted Entities & \bf Model &\bf Rewritten Outputs (Inputs to T5) & \bf Answers \\
\midrule
\multirow{9}{*}{
      \parbox[l]{1em}{
      \includegraphics[width=1.75in]{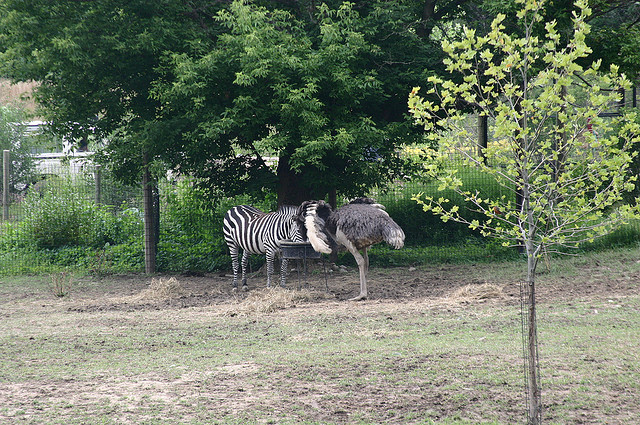}}}
& \multirow{9}{2.5cm}{what do these animals eat?}  
& \multirow{9}{2.8cm}{zebras, tree, park}  
& Gold & what do zebras eat? & grass, plants, leaves \\
\cline{4-6}
& & & Concat & what do these animals eat . zebras . tree . park & wild plants \\
\cline{4-6}
& & & Fine-Tuned T5 & - & hay \\
\cline{4-6}
& & & Model Agnostic & what do zebras eat? & leaves \\
\cline{4-6}
& & & Model Aware & what do zebras eat? & wild plants \\
\cline{4-6}
& & & MUTAN & - &  plant \\
\cline{4-6}
& & & MUTAN+AN & - & grass \\
\cline{4-6}
& & & BAN  & - & grass \\
\cline{4-6}
& & & BAN+AN & - & grass \\
\cline{4-6}
& & & QOnly & - & meat \\
\midrule
\multirow{15}{*}{
      \parbox[l]{1em}{
      \includegraphics[width=1.75in]{images/COCO_val2014_000000440539.jpg}}}
& \multirow{15}{2.5cm}{what famous founding father was known for his association with this object?\ \\}  
& \multirow{15}{*}{kite, sky, beach}  
& Gold & what famous founding father was known for his association with kite? & Benjamin Franklin \\
\cline{4-6}
& & & Concat & what famous founding father was known for his association with this object . kite . sky . beach & Benjamin Franklin \\
\cline{4-6}
& & & Fine-Tuned T5 & - & Benjamin Franklin \\
\cline{4-6}
& & & Model Agnostic & what famous founding father was known for his association with kite? & Benjamin Franklin \\
\cline{4-6}
& & & Model Aware & what famous founding father was known for his association with kite? & Benjamin Franklin \\
\cline{4-6}
& & & MUTAN & - &  kite \\
\cline{4-6}
& & & MUTAN+AN & - & shawn white \\
\cline{4-6}
& & & BAN  & - & beach boy \\
\cline{4-6}
& & & BAN+AN & - & Benjamin Franklin \\
\cline{4-6}
& & & QOnly & - & wright \\
\midrule
\multirow{15}{*}{
      \parbox[l]{1em}{
      \includegraphics[width=1.75in]{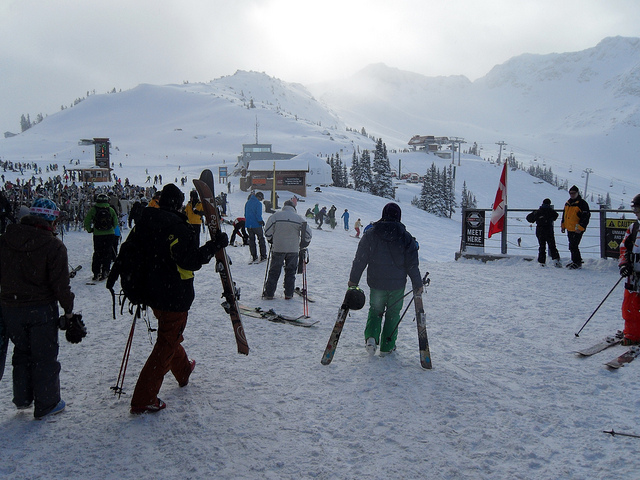}}}
& \multirow{15}{2.5cm}{what conditions are necessary for this sport?\ \\}  
& \multirow{15}{2.5cm}{skiing, dogsled, ski slope\ \\}  
& Gold & what conditions are necessary for skiing? & snow, snowfall \\
\cline{4-6}
& & & Concat & what conditions are necessary for this sport . skiing . dogsled . ski slope & snow \\
\cline{4-6}
& & & Fine-Tuned T5 & - & ice \\
\cline{4-6}
& & & Model Agnostic & what conditions are necessary for dogsled? \textit{(comment: bad rewrite but generates good answer)} & snow \\
\cline{4-6}
& & & Model Aware & what conditions are necessary for ski slope? \textit{(comment: bad rewrite but generates good answer)} & snowfall \\
\cline{4-6}
& & & MUTAN & - &  snow \\
\cline{4-6}
& & & MUTAN+AN & - & snow \\
\cline{4-6}
& & & BAN  & - & snow \\
\cline{4-6}
& & & BAN+AN & - & snow \\
\cline{4-6}
& & & QOnly & - & ski \\
\bottomrule
\end{tabular}}
    \caption{Examples outputs from different methods.}
    \label{tab:example}
\end{table*}

Out of 363 test data, Table \ref{tab:results} shows the results for different comparison systems using Exact match (EM), BERT similarity scores (BS), and human evaluation (HE).
We also present some anecdotal examples in Table \ref{tab:example}.
Except for the QOnly model, which has much lower accuracy, our methods achieve competitive results on exact match and BERT similarity scores in comparison with other VQA models.
Note that those VQA models are trained on significantly more human labeled data.
It is also worth noting that the exact match results contain many false negatives, because the metric assumes that the model generates the same answers as the gold standard answers.
Nevertheless, we can get other good answers such as the synonyms of the golden answers.
Similarly, BERT Similarity Score results can have false positives, because the answers that are of the same type as the golden answers will be encouraged but they might not be correct.
We thus report answer correctness assessment by humans as a reliable metric.
Based on human evaluation results, our model agnostic method and model aware methods seem to do better than the existing systems. We have several interesting findings:
\begin{itemize}
\item The difference between the exact match accuracy and human evaluation is larger for our methods than the comparison system because our methods are more likely to generate synonyms of the golden answers, which can be falsely ignored by the exact match metric.
T5 has a larger vocabulary size because it is trained on at least 409K questions across three datasets: Natural Questions \citep{kwiatkowski2019natural}, WebQuestions \citep{berant2013semantic} and TriviaQA \citep{joshi2017triviaqa}.
By contrast, those systems are only trained on 9,009 data from a single dataset OKVQA labeled by five MTurk workers. 
\item The fine-tuned T5 baseline is lower than the concatenated input across all three metrics, which shows that fine-tuning does not help on this task. 
\item Good rewrites usually generate good answers, but bad/grammatically incorrect rewrites (e.g., last row in Table \ref{tab:example}) may not lead to good answers. 
\end{itemize}

\section{Conclusion}
\label{sec:conclusion}
In this paper, we explore question rewrite strategies for knowledge-oriented VQA.
We contend that
for certain types of VQA question rewriting, a text-based question answering system can lower the needs for training data.
Unsupervised and weakly-supervised reinforcement learning-based question rewriting strategies are explored in this work.
We demonstrate effective VQA using an existing pre-trained \underline{text-based} QA model.
Experiments on these types of questions on a public dataset illustrate that this technique is competitive with other end-to-end VQA approaches.

For future work, we will investigate other types of text-based QA systems within this framework.
Since entity extraction and disambiguation are key steps, we plan to train an entity extractor with a larger label ontology for visual objects.
Additionally, we would like to investigate the impact of visual features, such as image position, bounding box information, relative positions of objects, etc. on the quality of the entity extractor.

\bibliography{aaai22}


\end{document}